\def\year{2022}\relax
\documentclass[letterpaper]{article} 
\usepackage{aaai22}  
\usepackage{times}  
\usepackage{helvet}  
\usepackage{courier}  
\usepackage[hyphens]{url}  
\usepackage{graphicx} 
\usepackage{natbib}  
\usepackage{caption} 
\DeclareCaptionStyle{ruled}{labelfont=normalfont,labelsep=colon,strut=off} 
\usepackage{algorithm}
\usepackage{algorithmic}
\usepackage{booktabs}
\usepackage{color}
\usepackage{bm}
\usepackage{amssymb}
\usepackage{amsmath}
\usepackage[switch]{lineno}  %
\usepackage{newfloat}
\usepackage{listings}
\floatstyle{ruled}
\newfloat{listing}{tb}{lst}{}
\floatname{listing}{Listing}

\title{One-shot Talking Face Generation from \\
Single-speaker Audio-Visual Correlation Learning}

\author{
    Suzhen Wang\textsuperscript{\rm 1}, 
    Lincheng Li\textsuperscript{\rm 1}, 
    Yu Ding\textsuperscript{\rm 1}\thanks{Corresponding author.}, 
    Xin Yu\textsuperscript{\rm 2} 
}
\affiliations{
    \textsuperscript{\rm 1} Netease Fuxi AI Lab \\
    \textsuperscript{\rm 2} University of Technology Sydney \\
    \{wangsuzhen, lilincheng, dingyu01\}@corp.netease.com \\
    xin.yu@uts.edu.au 
}

\usepackage{bibentry}

\begin{document}

\maketitle

\begin{abstract}
Audio-driven one-shot talking face generation methods are usually trained on video resources of various persons. However, their created videos often suffer unnatural mouth shapes and asynchronous lips because those methods struggle to learn a consistent speech style from different speakers. We observe that it would be much easier to learn a consistent speech style from a specific speaker, which leads to authentic mouth movements. Hence, we propose a novel one-shot talking face generation framework by exploring consistent correlations between audio and visual motions from a specific speaker and then transferring audio-driven motion fields to a reference image. Specifically, we develop an Audio-Visual Correlation Transformer (AVCT) that aims to infer talking motions represented by keypoint based dense motion fields from an input audio. In particular, considering audio may come from different identities in deployment, we incorporate phonemes to represent audio signals. In this manner, our AVCT can inherently generalize to audio spoken by other identities. Moreover, as face keypoints are used to represent speakers, AVCT is agnostic against appearances of the training speaker, and thus allows us to manipulate face images of different identities readily. Considering different face shapes lead to different motions, a motion field transfer module is exploited to reduce the audio-driven dense motion field gap between the training identity and the one-shot reference. Once we obtained the dense motion field of the reference image, we employ an image renderer to generate its talking face videos from an audio clip. Thanks to our learned consistent speaking style, our method generates authentic mouth shapes and vivid movements. Extensive experiments demonstrate that our synthesized videos outperform the state-of-the-art in terms of visual quality and lip-sync.

\end{abstract}

\begin{figure}[ht]
\centering
\includegraphics[width=0.5 \textwidth]{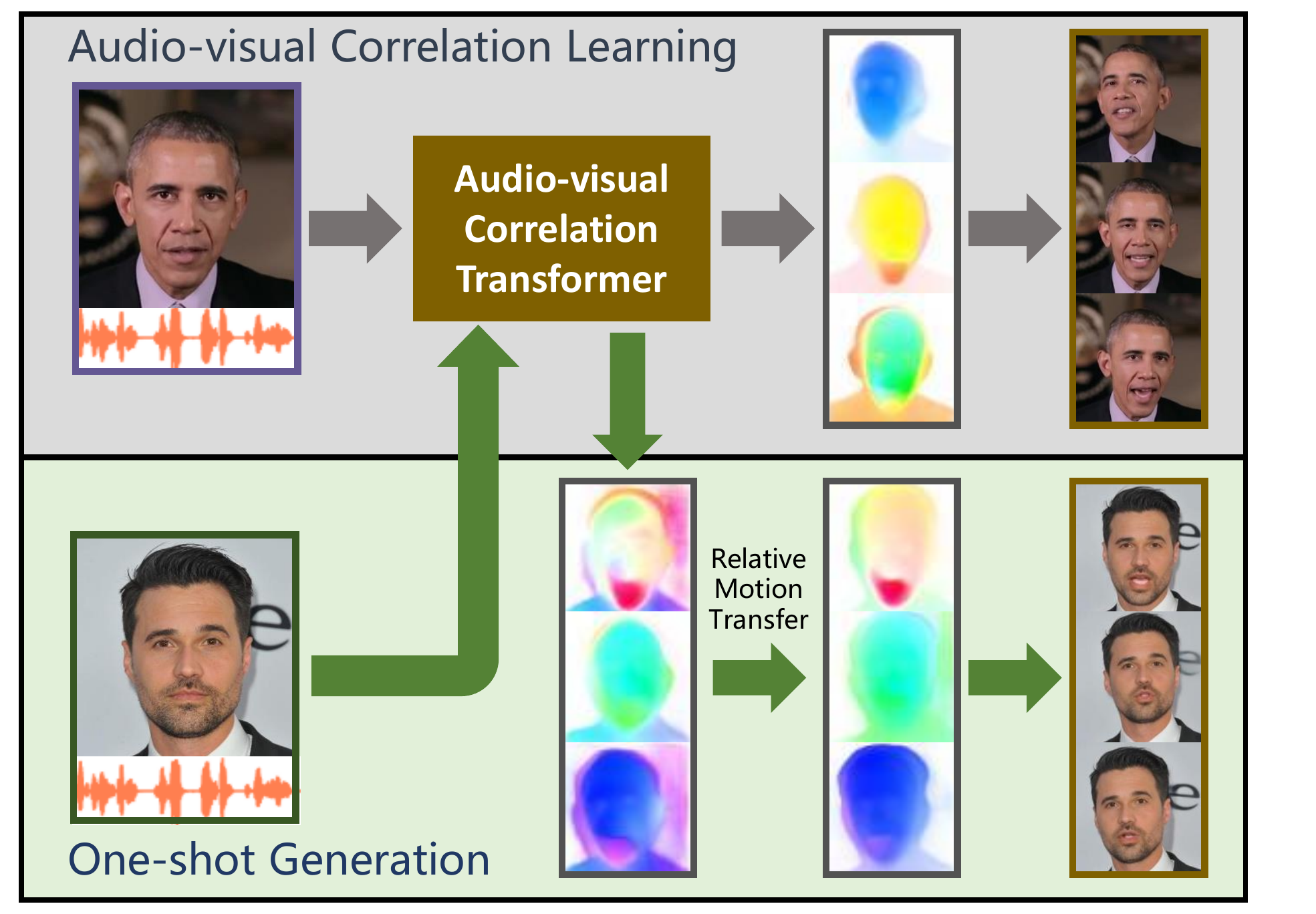}
\caption{Illustration of our proposed talking head generation framework. Our approach takes one-shot reference image as input and generates audio-driven talking faces with rhythmic head motions, natural mouth shapes and accurate lip synchronization. Although our audio-visual correlation model is trained on a specific speaker, our framework supports arbitrary one-shot reference image and voice as input and renders the photo-realistic talking face videos.}
\label{fig:illustration}
\end{figure}

\section{Introduction}
Synthesizing audio-driven photo-realistic portraits is of great importance to various applications, such as digital human animation \cite{ji2021audio,zhu2021deep}, visual dubbing in movies \cite{prajwal2020lip,ha2019marionette} and fast short video creation \cite{zhou2021pose,zeng2020realistic}. 
One-shot talking face generation methods are designed to animate video portraits for unseen speakers and voice.
When watching a synthetic talking head video, humans are mainly affected by three aspects: visual quality (clear and jitter-free), natural head motions, and synced lip movements. 
Existing one-shot methods  \cite{chen2019hierarchical,prajwal2020lip,zhou2020makelttalk,zhou2021pose,lah2021lipsync3d} are usually trained on video resources of various persons, and their results suffer unnatural lip shapes and bad lip-sync. 
This is mainly because their networks trained on multiple speech styles try to fit the common style among different identities while treating personalized variations as noise.
Therefore, it is very challenging to synthesize natural and synchronous lip movements for one-shot speakers.

We observe that it is much easier to learn a consistent speech style from a specific speaker.
Motivated by this, we propose a new one-shot talking face generation framework, which first learns a consistent speaking style from a single speaker and then animates vivid videos of arbitrary speakers with new speech audio from the learned style.
The key insight of the proposed method is to utilize the advantage of authentic lip movements and rhythmic head motions learned from an individual, and then to seek migration from an individual to multiple persons. 
Put differently, our method explores a consistent speech style between audio and visual motions from a specific speaker and then transfers audio-driven keypoint based motion fields to a reference image for talking face generation. 

Towards this goal, we firstly develop a speaker-independent Audio-Visual Correlation Transformer (AVCT) to obtain keypoint-base dense motion fields \cite{siarohin2019first} from audio signals.
To eliminate the timbre effect among different identities, we adopt phonemes to represent audio signals.
With the input phonemes and head poses, the encoder is expected to establish the latent pose-entangled audio-visual mapping. Considering vivid mouth movements are closely related to audio signals (e.g., the mouth amplitudes are affected by the fierce tones), we use the embedded acoustic features as the query of the decoder to modulate mouth shapes for more vivid lip movements. Moreover, due to the keypoint representation of a reference image, AVCT is agnostic against appearances of the training speaker, allowing us to manipulate face images regardless of different identities.
Furthermore, considering different faces have diverse shapes, these variations would lead to different facial motions. Thus, a relative motion transfer module~\cite{siarohin2019first} is employed to reduce the motion gap between the training identity and the one-shot reference. Once obtaining the dense motion field, we generate talking head videos by an image renderer.

Thanks to our learned consistent speaking style, our method is able to not only produce talking face videos of the training speaker on par with speaker-specific methods, but also animate vivid portrait videos for unseen speakers with more accurate lip synchronization and more natural mouth shapes than previous one-shot talking head approaches. 
Remarkably, our method can also address talking faces with translational and rotational head movements whereas prior arts usually handle rotational head motions.
Extensive experimental results on widely-used VoxCeleb2 and HDTF demonstrate the superiority of our proposed method.

In summary, our contributions are three-fold:
\begin{itemize}
\item We propose a new audio-driven one-shot talking face generation framework, which establishes the consistent audio-visual correlations from a specific speaker instead of learning from various speakers as in prior arts.
\item We design an audio-visual correlation transformer that takes phonemes and facial keypoint based motion field representations as input, thus allowing it to be easily extended to any other audio and identities. 

\item Although the audio-visual correlations are only learned from a specific speaker, our method is able to generate photo-realistic talking face videos with accurate lip synchronization, natural lip shapes and rhythmic head motions from a reference image and a new audio clip.

\end{itemize}

\section{Related Work}
Animating talking faces from audio or text has received more attention in the field of artificial intelligence. As there exists a considerable audio-visual gap, early works \cite{edwards2016jali,taylor2017deep,pham2017speech,karras2017audio,zhou2018visemenet,cudeiro2019capture} focus on driving animations of 3D face models. With the development of image generation~\cite{yu2016ultra,yu2019can,li2021super,yu2019semantic}, an increasing number of works have been proposed for 2D photo-realistic talking face generation. These methods can mainly be divided into two categories, speaker-specific methods and  speaker-arbitrary methods.

\subsection{Speaker-specific Talking Face Generation} 
For a given new speaker, speaker-specific methods retrain part or all of their models on the videos of that speaker. Most works \cite{suwajanakorn2017synthesizing,song2020everybody,yi2020audio,fried2019text,thies2020neural,li2021write, lah2021lipsync3d,ji2021audio,zhang20213d,zhang2021facial,lah2021lipsync3d} synthesize photo-realistic talking head videos guided by 3D face models.  \citet{suwajanakorn2017synthesizing} synthesize videos from audio in the region around the mouth.
Several methods \cite{thies2020neural,li2021write} consist of speaker-independent components relying on 3D face models and speak-specific rendering modules.  \citet{fried2019text} present a framework for text based video editing.
\citet{guo2021ad} propose the audio-driven neural radiance fields for talking head generation.


\begin{figure*}[ht]
\centering
\includegraphics[width=0.98\textwidth]{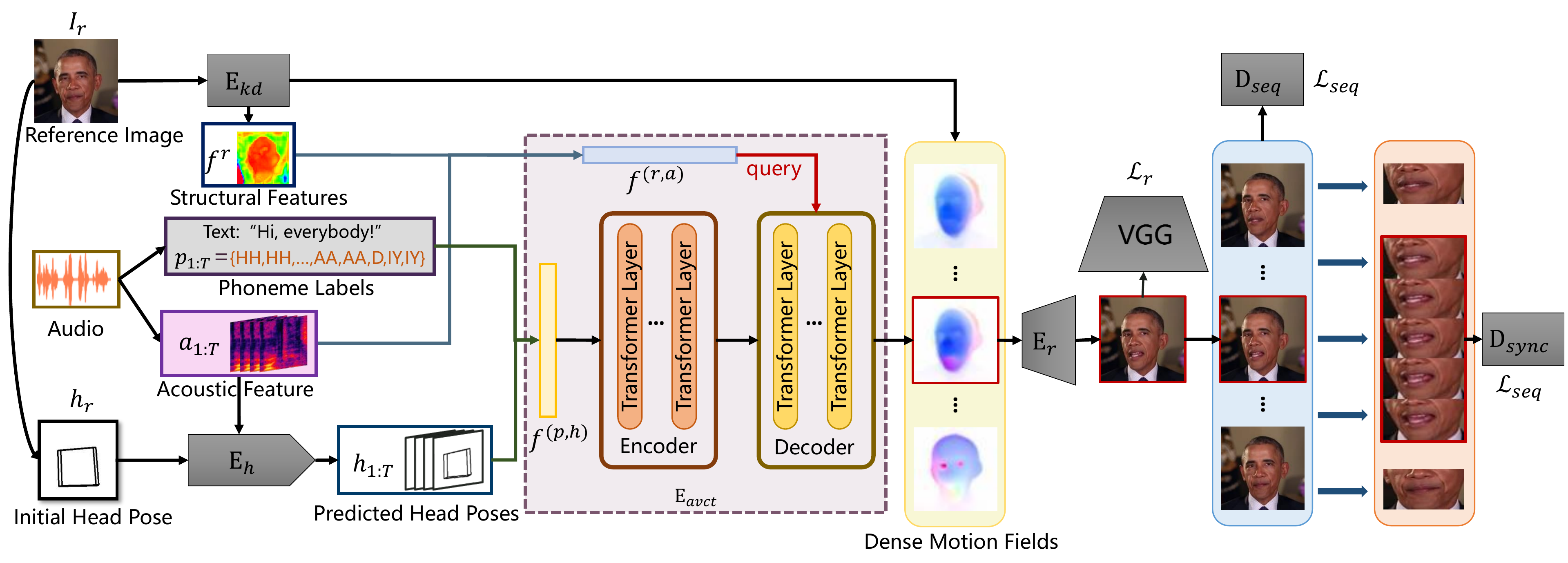}
\caption{Illustration of our pipeline. We first extract an initial pose $h_r$ from a reference image, and extract acoustic features $a_{1:T}$ and phoneme labels $p_{1:T}$ from the audio. The latent representation of keypoints of the reference image $f^r$ is extracted by the keypoint detector $\mathbf{E}_{kd}$. The head motion predictor $\mathbf{E}_h$ predicts the head motion sequence $h_{1:T}$ from the input $\{a_{1:T}, h_r\}$. Then $\{a_{1:T}, p_{1:T}, h_{1:T}, f^r\}$ are sent to the audio-visual correlation transformer $\mathbf{E}_{avct}$ 
to generate pose-aware keypoint-based dense motion fields. Finally, the image renderer $\mathbf{E}_r$ renders the output videos. We also use a temporal discriminator $\mathbf{D}_{seq}$ and a lip-sync discriminator $\mathbf{D}_{sync}$ to improve the temporal stability and lip synchronization respectively in training. Particularly, in the inference stage, the generated dense motions are refined with a relative motion transfer module.}
\label{fig:pipeline}
\end{figure*}

\subsection{Speaker-arbitrary Talking Face Generation} 
Speaker-arbitrary methods aim to build a single universal model for various subjects. Some works \cite{chung2017you,chen2018lip,song2018talking,zhou2019talking,chen2019hierarchical,vougioukas2019realistic,das2020speech} focus on learning a mapping from audio to the cropped faces, but their fixed poses and cropped faces in the videos are unnatural for human observations. Other works \cite{wiles2018x2face,chen2020talking,prajwal2020lip,zhou2020makelttalk,zhang2021flow,wang2021audio2head,zhou2021pose} try to create vivid results with natural head poses.  \citet{zhou2020makelttalk} predict 2D landmarks with the head pose, and then generate talking faces. \citet{chen2020talking} and \citet{zhang2021flow} use 3D face models to acquire landmarks and the dense flow respectively as their intermediate representations. 
\citet{zhou2021pose} propose a pose-controllable talking face generation method by implicitly modulating audio-visual representations. 
\citet{prajwal2020lip} only edit the lip-synced mouth regions of a reference image from audio. Although \citet{prajwal2020lip} and \citet{zhou2021pose} are able to generate videos with poses, their reference head poses are obtained from another videos rather than audio.  \citet{wang2021audio2head} employ keypoint-based dense motion fields as intermediate representations and achieve rhythmic head motions in generated videos, but their lip-sync exhibits artifacts. Among previous speaker-arbitrary works, only \citet{zhang2021flow} and  \citet{wang2021audio2head} create videos with translational and rotational head movements while keeping background still in generated videos. However, since all these methods are trained on the corpus of multiple speakers, they often struggle to learn a consistent speaking style, and their results suffer unnatural mouth shapes.

\section{Proposed Method}
We propose a new talking face generation framework to make audio-driven portrait videos for arbitrary speakers by learning audio-visual correlations on a specific speaker. Giving a reference image $I_r$ and an audio clip $A$, our method creates talking face images $\bm{y}$=$I_{1:T}$. The whole pipeline is shown in Figure~\ref{fig:pipeline}. Our pipeline consists of four modules: (1) a head motion predictor $\mathbf{E}_h$ estimates the head motion sequence $h_{1:T}$ ($h_i \in \mathbb{R}^6$ includes the 3D rotation and the 3D translation) from audio; (2) a keypoint detector $\mathbf{E}_{kd}$ extracts initial keypoints from the reference image; (3) an Audio-visual Correlation Transformer (AVCT) $\mathbf{E}_{avct}$ maps audio signals to keypoint-based dense motion fields; and (4) an image renderer $\mathbf{E}_r$ produces output images from the dense motion fields.
we adopt the architectures of $\mathbf{E}_{kd}$ and $\mathbf{E}_r$ as in FOMM \cite{siarohin2019first}.

We extract the audio channels from the training videos and transform them into audio features and phonemes as pre-processing. To be consistent with videos at 25 fps, we extract acoustic features $a_i \in \mathbb{R}^{4 \times 41 }$ and one phoneme label $p_i \in \mathbb{R}$ per 40ms. 
The acoustic features include Mel Frequency Cepstrum Coefficients (MFCC), Mel-filterbank energy features (FBANK), fundamental frequency and voice flag. The phoneme is extracted by a speech recognition tool\footnote{https://cmusphinx.github.io/wiki/phonemerecognition/}.

\subsection{Audio-visual Correlation Transformer (AVCT)}
The core of the proposed method is to build accurate audio-visual correlations which can be extended to any other audio and identities. 
Such correlations are learned via a speaker-independent audio-visual correlation transformer. 
Considering the high temporal coherence, $\mathbf{E}_{avct}$ takes the assembled features in a sliding window as input. Specifically, for the $i$-th frame, $\mathbf{E}_{avct}$ takes the paired conditioning input  $c_i = \{f_r$, $a_{i-n:i+n}$, $h_{i-n:i+n}$, $p_{i-n:i+n}\}$ and outputs the keypoints $k_i \in \mathbb{R}^{N \times 2}$ and their corresponding Jacobian $j_i \in \mathbb{R}^{N \times 2 \times 2}$. $f^r$ is the latent representation of keypoints from the reference image $I_r$ through the keypoint detector $\mathbf{E}_{kd}$.
$n$ indicates the window length and is set to 5 in our experiments. N is the number of keypoints and is set to 10. The paired $(k_i,j_i)$ represents the dense motion field \cite{siarohin2019first}. The head motions $h_{1:T}$ are extracted by OpenFace \cite{baltrusaitis2018openface}.

The proposed AVCT is able to aggregate dynamic audio-visual information within the temporal window, thus creating more accurate lip movements.
To model the correlations among different modalities, we employ Transformer \cite{vaswani2017attention} as the backbone of AVCT due to its powerful attention mechanism.
For a better extension to any other audio and identities, we carefully design the Encoder and the Decoder of $E_{avct}$ as follows.

\subsubsection{Encoder.}
We employ the phoneme labels as input instead of acoustics features to bridge the timbre gap between the specific speaker and arbitrary ones. 
We establish a latent pose-entangled audio-visual mapping by encoding the input phonemes $p_{i-n:i+n}$ and poses $h_{i-n:i+n}$ in the encoder.
The attentions between sequential frames ${2n}$+${1}$ allow for obtaining the refined latent mouth motion representation at frame $i$.
Specifically, we employ a 256-dimension word embedding \cite{levy2014neural} to represent the phoneme label, then reshape and upsample it as $f^p_i \in \mathbb{R}^{1\times64\times64}$. $h_i$ is converted to the projected binary image $f^h_i \in \mathbb{R}^{1\times64\times64}$ as in \cite{wang2021audio2head}.
Then, the concatenated features $\{f^p_i,f^h_i\}$ are fed into a residual convolution network consisting of 5 $2\times$ downsampling ResNet blocks \cite{he2016deep} in order to obtain the assembled feature $f^{(p,h)}_i \in \mathbb{R}^{1 \times 512}$. 

The input to the encoder is the sequential features $f^{(p,h)} \in \mathbb{R}^{(2n+1) \times 512}$ by concatenating all the frame features along the temporal dimension.
Since the architecture of the transformer is permutation-invariant, we supplement $f^{(p,h)}$ with fixed positional encoding \cite{vaswani2017attention}.

\begin{table*}[!htbp] 
\centering
\caption{The quantitative results on HDTF and VoxCeleb2.}
\begin{tabular}{ccccccccccc}
\toprule  
&\multicolumn{5}{c}{HDTF}&\multicolumn{5}{c}{VoxCeleb2} \\
\cmidrule(r){2-6}  \cmidrule(r){7-11}
Method & FID$\downarrow$ & CPBD$\uparrow$ & LMD$\downarrow$ & AVOff($\rightarrow$0) & AVConf$\uparrow$ & FID$\downarrow$ & CPBD$\uparrow$ & LMD$\downarrow$ & AVOff($\rightarrow$0) & AVConf$\uparrow$ \\
\midrule

Wav2Lip       & 0.180 & \textbf{0.790} & 0.289 & -2.92 & 6.97 & 0.203 & 0.541 & 0.273 & -2.92 & 6.65  \\
MakeitTalk    & 0.210 & 0.694 & 0.546 & -2.93 & 4.87 & 0.230 & 0.550 & 0.482 & -2.83 & 4.38  \\
Audio2Head    & 0.176 & 0.732 & 0.483 & \textbf{0.33}  & 3.90 & 0.224 & 0.532 & 0.314 & 0.50  & 2.47  \\
FGTF          & 0.187 & 0.738 & 0.387 & 1.05  & 4.24 & 0.212  & 0.559 & 0.283 & 0.491 & 4.54 \\
PC-AVS        & 0.238 & 0.725 & 0.318 & -3.00 & \textbf{7.18} & 0.276 & 0.514 & \textbf{0.251} & -3.00  & 6.83  \\
Ground Truth  & 0     & 0.827 &     0 & 0.15  & 8.58 & 0     & 0.612 & 0     & -2.33  & 7.16  \\
\cmidrule(r){1-11}
\textbf{Ours} & \textbf{0.172} & 0.751 & \textbf{0.271} & \textbf{-0.33} & 7.09 & \textbf{0.194} & \textbf{0.564} & 0.252 & \textbf{-0.08} & \textbf{6.98} \\
\bottomrule 
\end{tabular}
\label{table:quantitive_evaluation}
\end{table*}

\begin{figure*}[ht]
\centering
\includegraphics[width=1.0 \textwidth]{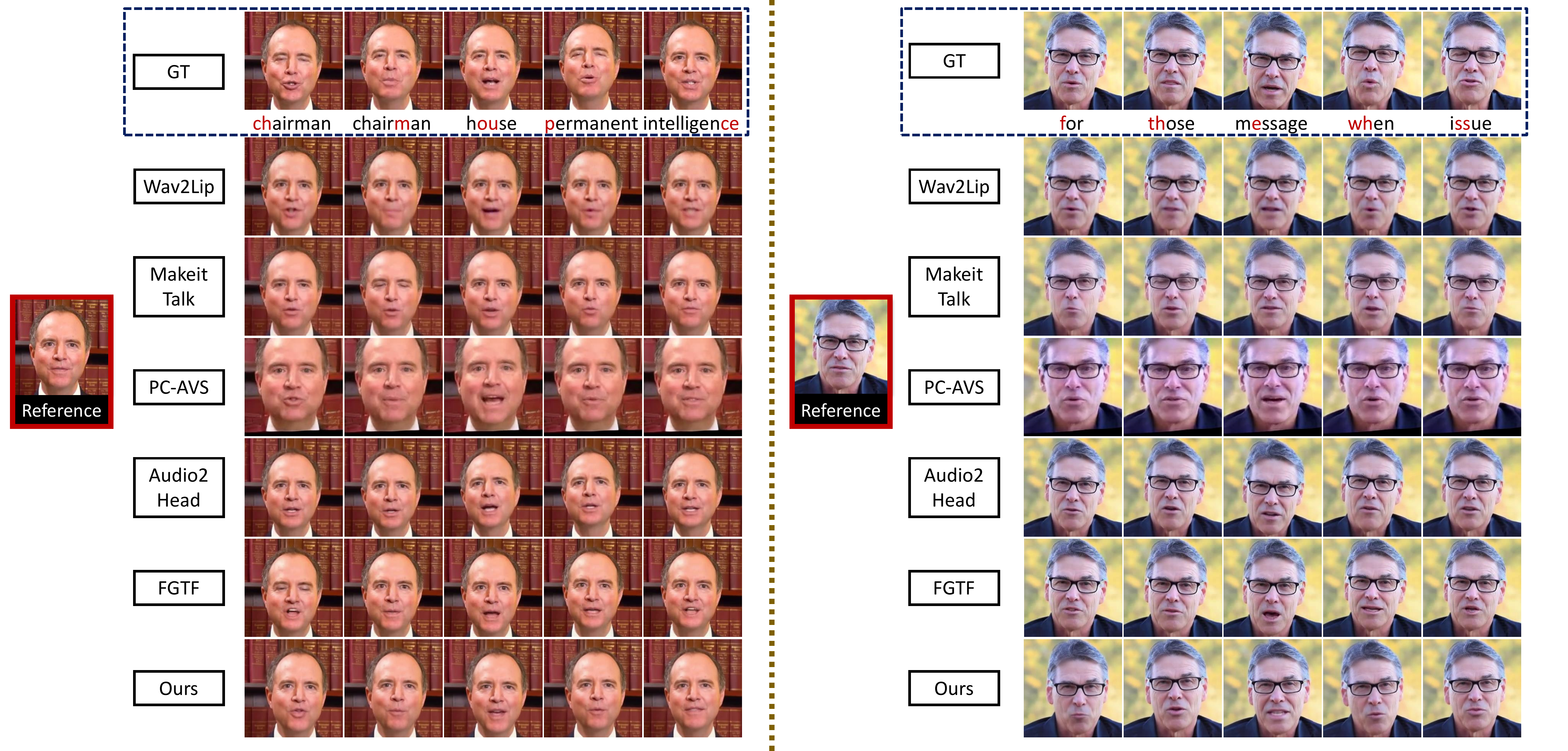}
\caption{Comparisons with the speaker-arbitrary methods. We select the frames that pronounce the same phonemes marked in red. Please see our demo videos for more details.}
\label{fig:with_arbitrary}
\end{figure*}

\subsubsection{Decoder.}
Practically, the mouth amplitude is affected by the loudness and energy of the audio in addition to the phoneme. To create more subtle mouth movements, we employ the acoustics features in the decoding phase for capturing energy changes. We extract audio features $f^a_i \in \mathbb{R}^{32 \times 64 \times 64} $ from $a_i$ using an upsampling convolution network. 
To reduce the dependency on the identities, we cannot directly take the reference image as input but employ the latent representation, $f^r \in \mathbb{R}^{32 \times 64 \times 64} $, of the keypoints of the reference image $I_r$. $f^r$ is extracted from the pretrained keypoint detector $\mathbb{E}_{kd}$. It mainly retains the pose-based structural information of body, face and background, weakening identity-related information. Such initial structural information dominates the low-frequency holistic layout in the generated dense motion fields.

$f^r$ is repeated by $2n$+$1$ times. Then the concatenation of $f^r_i$ and $f^a_i$ is fed into another residual convolutional network to obtain the embedding $f^{(r,a)}_i$. Similarly, we acquire the features $f^{(r,a)} \in \mathbb{R}^{(2n+1) \times 512}$ by concatenation, and supplement it with positional encodings. $f^{(r,a)}$ is used as the initial query of the decoder to modulate the layout of the body, head and background, and to refine the subtle mouth shape. 
Following the standard transformer, the decoder creates $2n$+$1$ embeddings. Only the $i$-th embedding is taken and projected to keypoints $k_i$ and Jacobians $j_i$ with two different linear projections.


\subsection{Batched Sequential Training}
Since AVCT generates the dense motion fields of each frame individually, we develop a batched sequential training strategy to improve the temporal consistency.
The training samples on each batch consist of the T successive conditional inputs $c_{i:T}$ of T successive images from the same video. Then, we generate image sequence $\hat{I}_{1:T}$ in parallel in each batch. This design allows us to apply constraints on the image sequence within each batch rather than on single images. We call the above strategy as Batched Sequential Training (BST). 
In addition to the pixel loss on each frame image,
the sequential constraint is imposed by a temporal discriminator $\mathbf{D}_{seq}$. 
Besides, as the common pixel reconstruction loss is insufficient to supervise the lip-sync, we employ another lip-sync discriminator $\mathbf{D}_{sync}$ to improve the lip-sync.


\subsubsection{Temporal Discriminator.} $\mathbf{D}_{seq}$ follows the structure of PatchGAN \cite{goodfellow2014generative,isola2017image,yu2017face,yu2017hallucinating,yu2018face}. We stack the $T$ successive image frames along the channel dimension as the input of $\mathbf{D}_{seq}$. $\mathbf{D}_{seq}$ tries to distinguish whether the input is natural or synthetic. $\mathbf{D}_{seq}$ and $\mathbf{E}_{avct}$ are learned jointly with the generative-adversarial learning.

\subsubsection{Lip-sync Discriminator.} $\mathbf{D}_{sync}$ employs the structure of SyncNet \cite{chung2016out} in Wav2Lip \cite{prajwal2020lip}. $\mathbf{D}_{sync}$ is trained to discriminate the synchronization between audio and video by randomly sampling an audio window that is either synchronous or asynchronous with a video window. The discriminated frame lies in the middle of the window, and the window size is set to 5. $\mathbf{D}_{sync}$ computes the visual embedding $e_v$ from an image encoder and the audio embedding $e_a$ from an audio encoder. We adopt the cosine-similarity to indicate the probability whether $e_v$ and $e_a$ are synchronous:
\begin{equation}
    P_{sync} =  \frac{e_v \cdot e_a}{max({\| e_v \|}_2 \cdot {\| e_a \|}_2, \epsilon)}.
\end{equation}

\subsubsection{Loss Function.} Based on the batched sequential training, the loss function for each batch image sequence is defined as:
\begin{equation}
\begin{split}
\mathcal{L}_{total} =\mathcal{L}_{seq}(\hat{I}_{1:T}) + \frac{\lambda_{sync}}{T-4} \sum_{i=3}^{T-2}\mathcal{L}_{sync}(\hat{I}^{crop}_{i-2:i+2}) + \\
+ \frac{1}{T} \sum_{i=1}^T(\lambda_{v}\mathcal{L}_{vgg}^{mul}(\hat{I}_i,I_i)+\lambda_{eq}^P \mathcal{L}_{eq}^{K}(\hat{k}_i) + \lambda_{eq}^J \mathcal{L}_{eq}^{J}(\hat{j}_i))
\end{split} 
\end{equation}
where $\mathcal{L}_{seq}$ is the GAN loss of $\mathbf{D}_{seq}$. $\mathcal{L}_{sync}$ is the lip-sync loss from the pretrained $\mathbf{D}_{sync}$ and is defined as $-log(P_{sync})$. Note that $\hat{I}^{crop}$ means the cropped mouth area and that we ignore the boundary frames to fit the temporal input of $\mathbf{D}_{sync}$. We crop the mouth region in each training iteration dynamically according to the detected bounding boxes of real videos. $\mathcal{L}_{vgg}^{mul}$ is the multi-layer perceptual loss that relies on a pretrained VGG network. $\mathcal{L}_{eq}^{K}$ and $\mathcal{L}_{eq}^{J}$ are the equivariance constraint loss \cite{siarohin2019first} to ensure the consistency of estimated keypoints and jacobians. In our experiments, $T$ is set to 24 (on RTX 3090), $\lambda_{sync}$,$\lambda_{v}$,$\lambda_{eq}^{P}$ and $\lambda_{eq}^J$ are set to 10,1,10,10 respectively.

\subsection{Head Motion Predictor}
The head motion predictor $\mathbf{E}_h$, developed to generates $h_{1:T}$ in the inference stage, is also trained on the specific speaker.
$\mathbf{E}_h$ adopts the network structure of the head motion predictor of Audio2Head \cite{wang2021audio2head} but has two differences. 
First, instead of being trained on a large number of identities, $\mathbf{E}_h$ is trained on a specific speaker. Therefore, in order to avoid overfitting to the appearance of the specific speaker, we replace the input reference image of Audio2Head with the projected binary pose image. 
Secondly, for the convenience of the relative motion transfer (see below), the starting point of the predicted head pose sequence should be the same as the head pose of the reference image. Hence, we add an $\textbf{L}1$ loss term between the head poses of the first predicted frame and the reference image to the primitive loss function when training $\mathbf{E}_h$.

\subsection{Relative Motion Transfer}
As the generated motion fields are inevitably entangled with the specific speaker, we adopt the relative motion transfer \cite{siarohin2019first} in the inference stage to reduce the motion gap between the training identity and the one-shot reference image. We transfer the relative motion between $(\hat{k}_1,\hat{j}_1)$ and $(\hat{k}_{1:T},\hat{j}_{1:T})$ to $(k_r,j_r)$. $(k_r,j_r)$ are detected from the reference image. This operation is defined by:
\begin{equation}
    \hat{k}_i' = \hat{k}_i-\hat{k}_1 + k_r, \quad \hat{j}_i' = \hat{j}_i \hat{j}_1^{-1}  j_r.
\end{equation}
Then, we use $(\hat{k}'_{1:T},\hat{j}'_{1:T})$ to render the output videos. 

\section{Experiments}
\subsubsection{Dataset.}
We collect Obama Weekly Address videos from Youtube. 
Since we aim to generate talking face videos with translational and rotational head movements, we crop and resize the original videos to 256×256 pixels as in FOMM \cite{siarohin2019first} without aligning speakers' noses across frames. 
After processing all the crowdsourcing videos, we obtain 20 hours of videos of Obama. 
Although our audio-visual correlation transformer is only trained on Obama, we employ two in-the-wild audio-visual datasets to evaluate our method, HDTF \cite{zhang2021flow} and VoxCeleb2 \cite{chung2018voxceleb2}.

\subsubsection{Evaluation Metrics.} We conduct quantitative evaluations on several metrics that have been wildly used in previous methods. As the generated videos have different head motions from ground truth, we use the Fréchet Inception Distance (FID) \cite{heusel2017gans} and the Cumulative Probability of Blur Detection (CPBD) \cite{narvekar2009no} to evaluate the image quality. To evaluate the mouth shape and lip synchronization, we adopt the Landmark Distance (LMD) \cite{chen2019hierarchical} around mouths and audio-visual metrics (AVOff and AVConf) proposed in SyncNet \cite{chung2016out}. Note that we calculate the normalized relative landmark distance instead of the absolute landmark distance to avoid the influence of both head poses and image resolution.

\subsubsection{Implementation Details.}
All models are implemented by PyTorch, and we adopt Adam \cite{kingma2014adam} optimizer for all experiments.
Before training our AVCT $\mathbf{E}_{avct}$, we train the keypoint detector and image renderer on the combination of VoxCeleb \cite{nagrani2017voxceleb} and Obama datasets to obtain the pretrained $\mathbf{E}_{kd}$ and $\mathbf{E}_{r}$. 
$\mathbf{D}_{sync}$ is trained with a fixed learning rate 1e-4 on videos of Obama. 
$\mathbf{E}_{kd}$, $\mathbf{E}_{r}$ and $\mathbf{D}_{sync}$ are frozen when training $\mathbf{E}_{avct}$ on the videos of Obama. $\mathbf{E}_{avct}$ is trained on 4 GPU for about 5 days using the batched sequential training mechanism, with an initial learning rate of 2e-5 and a weight decay of 2e-7. $\mathbf{E}_{h}$ is trained on a single GPU for about 12 hours with a learning rate of 1e-4.

\begin{figure}[t]
\centering
\includegraphics[width=0.49\textwidth]{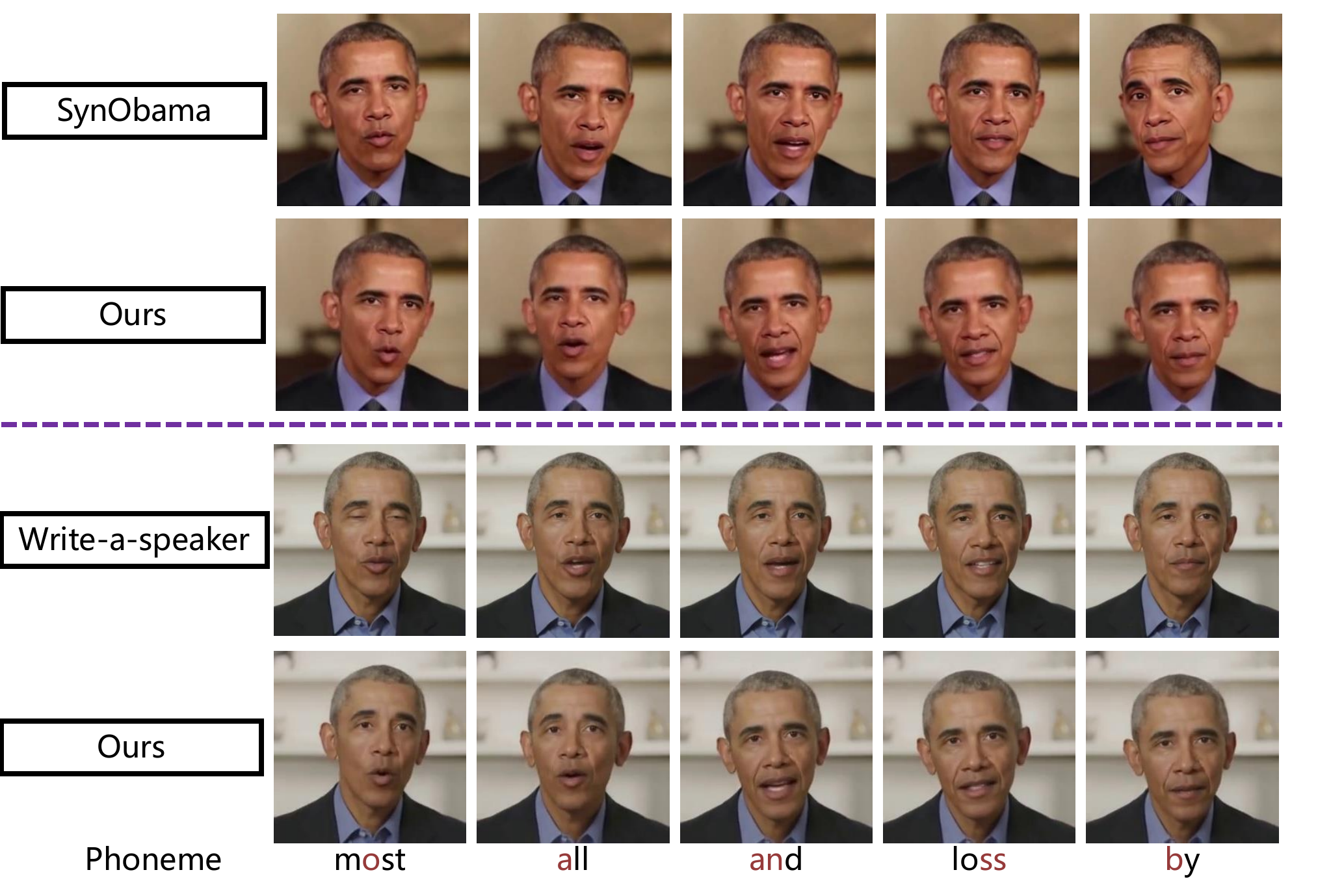}
\caption{Comparisons with the speaker-specific methods. We select the frames that pronounce the same phonemes.}
\label{fig:with_specific}
\end{figure}

\begin{table}[t]
\small
\centering
\caption{Results of ablation study on HDTF.}
\setlength{\tabcolsep}{1.4mm}{
\begin{tabular}{cccccc}
\toprule  
Method & FID$\downarrow$ & CPBD$\uparrow$ & LMD$\downarrow$ & AVOff($\rightarrow$0) & AVConf$\uparrow$ \\
\midrule

w/o Pho                  & 0.155 & 0.740 & 0.376 & -0.503 & 6.13 \\
w/o Aud                  & 0.151 & \textbf{0.747} & 0.293 & -0.538 & 6.59 \\
w/o Dec$_{\rm kp}$                  & 0.184 & 0.707 & 0.418 & -0.308 & 5.24 \\
w/o Kp           & 0.210 & 0.704 & 0.461 & \textbf{-0.170} & 4.92  \\
w/o BST                  & 0.151 &\textbf{0.747} & 0.388 & -0.743 & 4.55 \\
w/o $\mathbf{D}_{sync}$  & 0.150 & 0.741 & 0.385 & -0.385 & 5.53\\
\cmidrule(r){1-6}
\textbf{Full}            & \textbf{0.148} & 0.740 & \textbf{0.274} & -0.417 & \textbf{7.17}\\
\bottomrule 
\end{tabular}}
\label{table:Ablation_study}
\end{table}

\subsection{Quantitative Evaluation}
We compare our method with recent state-of-the-art methods, including Wav2Lip \cite{prajwal2020lip}, MakeitTalk \cite{zhou2020makelttalk}, Audio2Head \cite{wang2021audio2head}, FGTF \cite{zhang2021flow}, and PC-AVS \cite{zhou2021pose}. The samples of each method are generated using their released codes with the same reference image and audio. The reference images are specially cropped to fit the input of PC-AVS. 
Since Wav2Lip and PC-AVS cannot obtain head poses from audio, the head poses are fixed in their generated videos. For other methods, the head poses are controlled separately by each method. The quantitative results are reported in Table \ref{table:quantitive_evaluation}. 
Our method achieves the best performance under most of evaluation metrics on HDTF and VoxCeleb2. As Wav2Lip only edits the mouth regions and keeps most parts of the reference image unchanged, it reaches the highest CPBD score on HDTF, but their synthetic mouth areas are noticeably blurry, as visible in Figure \ref{fig:with_arbitrary}. These results validate that our method achieves high-quality videos, even though our audio-visual correlation transformer is learned from a specific speaker. 

\begin{table*}[ht] 
\centering
\caption{Results of user study. Participants rate each video from 1 to 5. Large scores indicate better visual quality. Here, the average scores across 21 videos are reported.}
\begin{tabular}{cccccccc}
\toprule  
Method & Wav2Lip & MakeitTalk & Audio2Head & FGTF & PC-AVS & Ground Truth & Ours \\
\midrule
Lip Sync Quality & 3.80 & 2.65 & 2.07 & 2.43 & 3.74 & 4.96 & \textbf{4.32} \\
Head Movement Naturalness & 1.21 & 2.04 & 3.79 & 3.74 & 1.79 & 4.89 & \textbf{4.11} \\
Video Realness & 1.68 & 1.70 & 3.21 & 3.08 & 2.14 & 4.89 & \textbf{3.86} \\
\bottomrule 
\end{tabular}
\label{table:user_study}
\end{table*}

\begin{figure}
\centering
\includegraphics[width=0.5\textwidth]{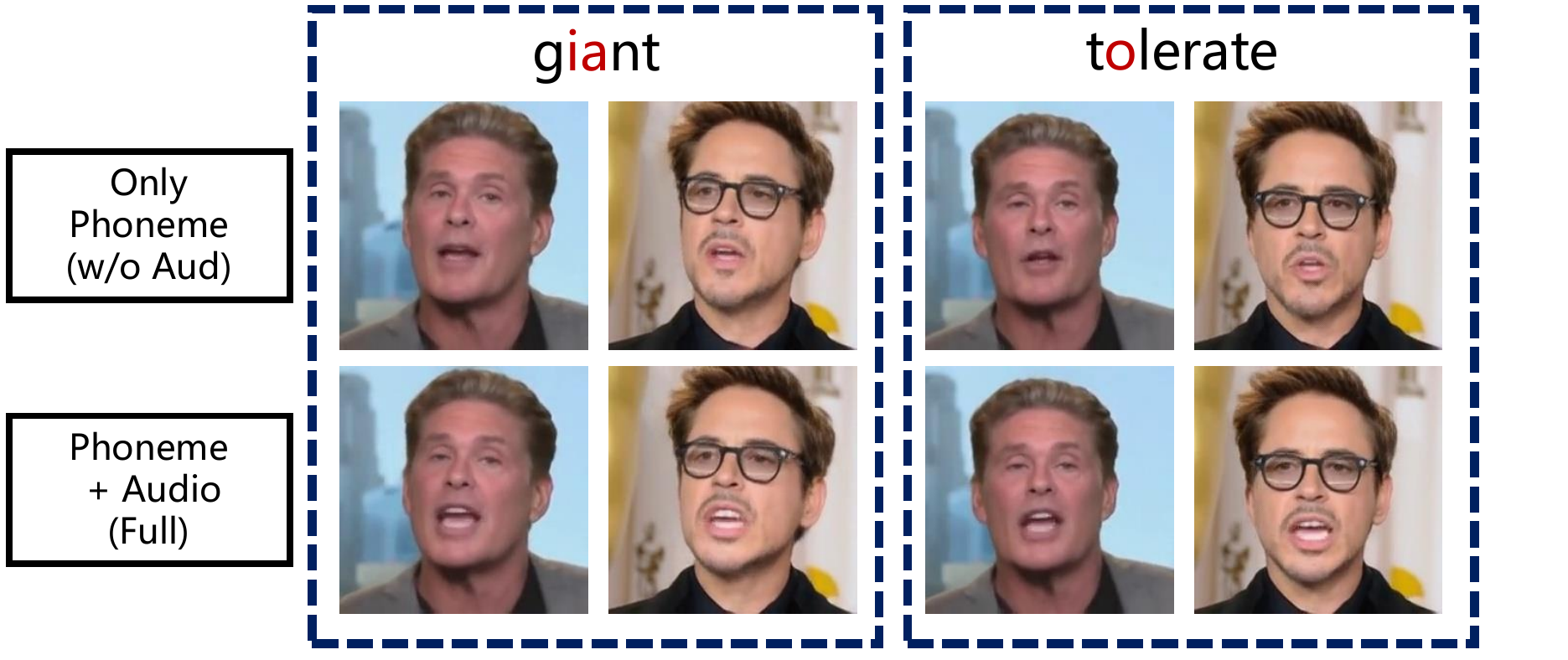}
\caption{Samples are generated from different representations of the same audio clip, \emph{i.e.}, phoneme features and the combination of phoneme and audio features. 
Here, the mouth shapes are further controlled by the intensity of the pronunciation in our method. 
}
\label{fig:ablation_audio}
\end{figure}

\begin{figure}
\centering
\includegraphics[width=0.5\textwidth]{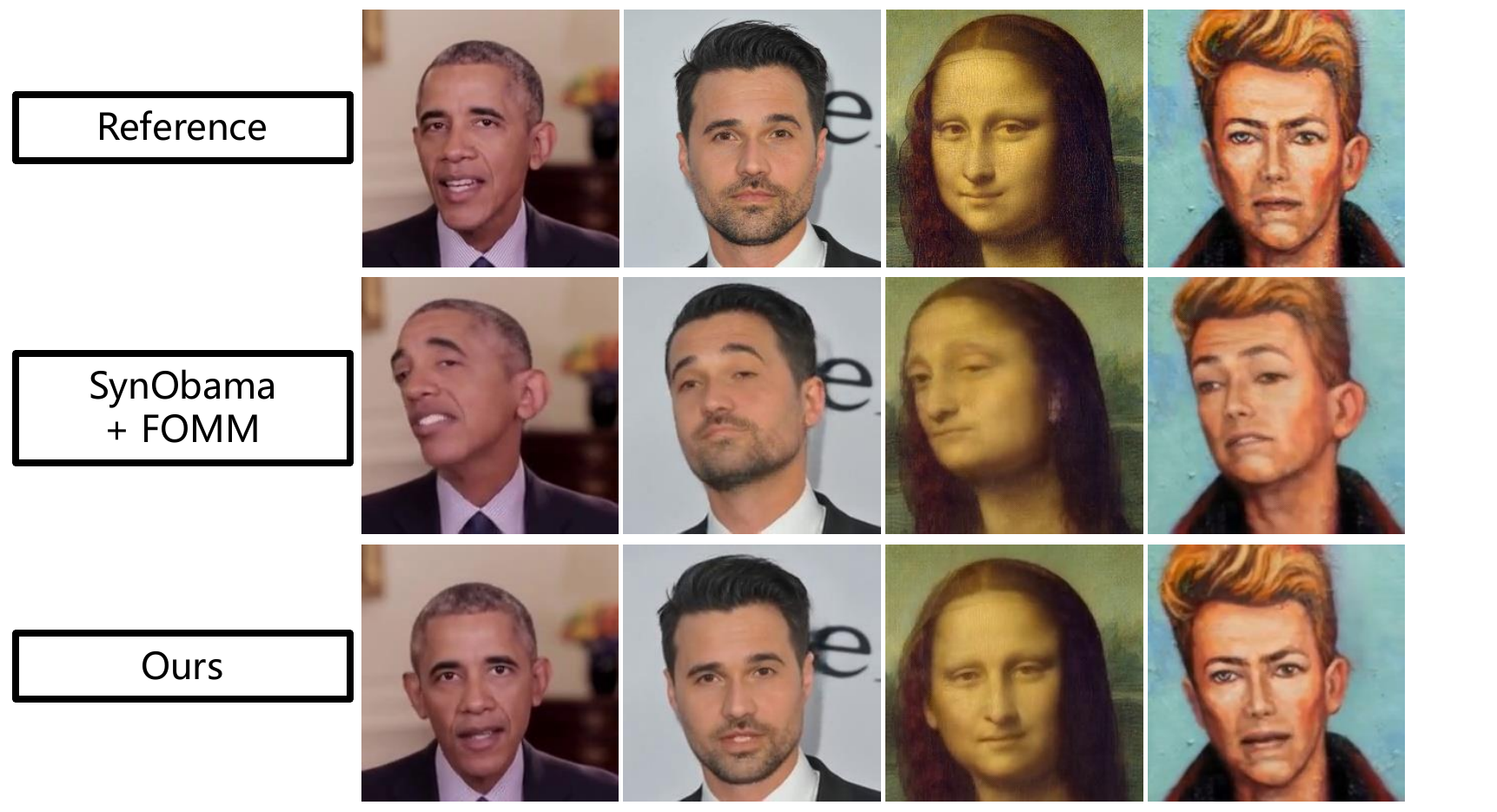}
\caption{Compare our results with videos created by the combination of SynObama and FOMM. The combinated method is sensitive to the initial pose, while our method preserves the authentic head motions.}
\label{fig:ablation_trans}
\end{figure}

\subsection{Qualitative Evaluation}

\subsubsection{Comparisons with Speaker-arbitrary Methods.} We first compare our method with speaker-arbitrary (one-shot) methods qualitatively. The results are shown in Figure \ref{fig:with_arbitrary}. Among all methods, our method creates the most accurate mouth shape and preserves the best identity (see our demo video for more clearly comparisons). Only Wav2Lip and PC-AVS achieve similar lip-sync 
to ours, but their mouth shapes look mechanical and unnatural because these methods struggle to produce consistent talking styles. Moreover, both of them cannot obtain head poses from audio and PC-AVS can only deal with aligned faces. Notably, PC-AVS alters the identity information of the reference image in Figure~\ref{fig:with_arbitrary}. While MakeitTalk creates subtle head motions, its results are obviously out of sync. Audio2Head, FGTF and our method are able to produce talking face videos with moving head poses. 
However, Audio2Head still suffer inferior lip-sync while our method produce authentic lip-sync. In addition, FGTF produces distorted mouth shapes in its results while our generated mouths look very natural.


\subsubsection{Comparisons with Speaker-specific Methods.} We compare our method with two speaker-specific methods, \emph{i.e.}, SynObama \cite{suwajanakorn2017synthesizing} and Write-a-speaker \cite{li2021write}. We first extract the reference image and audio from their demo videos and then generate our results. 
The comparisons are shown in Figure \ref{fig:with_specific}. 
As shown in Figure \ref{fig:with_specific}, our method synthesizes comparable videos of Obama with the methods that are customized for Obama.
SynObama synthesizes the region around the mouth from audio, and uses compositing techniques to borrow the rest regions from real videos. 
This composition sometimes results in visible artifacts around the mouth. 
Our method is an end-to-end approach without requiring additional editing, and achieves accurate mouth shapes than Write-a-speaker. 
Please see our supplementary video for more details.

\subsection{Ablation Study}
To evaluate the effectiveness of each component in our framework, we conduct an ablation study with 7 variants: 
(1) remove phonemes from the encoder (\textbf{w/o Pho}), 
(2) remove audio features from the decoder (\textbf{w/o Aud}), 
(3) remove the keypoint features in the decoder (\textbf{w/o Dec$_\textbf{kp}$}), 
(4) replace the extracted keypoint features from $\mathbf{E}_{kd}$ with the reference image (\textbf{w/o Kp}), 
(5) remove the temporal discriminator $\mathbf{D}_{seq}$ and lip-sync discriminator $\mathbf{D}_{sync}$ (\textbf{w/o BST}), 
(6) only remove $\mathbf{D}_{sync}$ (\textbf{w/o $\mathbf{D}_{sync}$}), and (7) our full model (\textbf{Full}). 
For evaluation, we replace generated head motions with poses extracted from real videos to create the samples. The numerical results on HDTF are shown in Table \ref{table:Ablation_study}. As all the variants employ the same pretrained image renderer, most of them achieve similar FID and CPBD scores. However, it can be seen that the image quality drops dramatically when removing the reference image or replacing the keypoint features with the reference image. Without the supervision of $\mathbf{D}_{seq}$ and $\mathbf{D}_{sync}$, the results show poor temporal stability and bad lip synchronization. The model \textbf{w/o Pho} fails to extend to the unseen timbre by removing the input of phonemes, producing out-of-sync videos. 
In Table \ref{table:Ablation_study}, the model \textbf{w/o Aud} obtains good quantitative results without using audio features. However, as seen in Figure \ref{fig:ablation_audio}, the audio features indeed affect the vivid mouth movements.

We also conduct another ablation study to evaluate the superiority of the proposed method. We compare our method with the combination of a speaker-specific work (\emph{i.e.}, SynObama) and a expression transfer work (\emph{i.e.}, FOMM). Specifically, we transfer the expressions in videos created by SynObama to the reference image using FOMM. The results are shown in Figure \ref{fig:ablation_trans}.
Since FOMM requires the initial poses of the reference image and one-shot to be similar while the pose of the first frame from SynObama is possibly different from that of the reference one, 
the combination of SynObama and FOMM would lead to inferior results.
In contrast, our model is able to preserve the initial pose and thus generates satisfactory facial motions.


\subsection{User Study}
We conduct a user study of 19 volunteers base on their subjective perception of talking head videos. We create 3 videos for each method with the same input and obtain $21$ videos in total. We adopt the questions used in the user study of PC-AVS  \cite{zhou2021pose}, and participants are asked to give their ratings (1-5) of each video on three questions: (1) the Lip sync quality, (2) the naturalness of head movements, and (3) the realness of results. The mean scores are listed in Table \ref{table:user_study}. Note that we do not offer reference poses for Wav2Lip and PC-AVS, so their scores on head movements and video realness are reasonably low. Our method outperforms competing methods in all the aspects, demonstrating the effectiveness of our method.

\section{Conclusion}

In this paper, we propose a novel framework for one-shot talking face generation from audio. 
Particularly, our method learns consistent audio-visual correlations from a single speaker, and then transfers the talking styles to different subjects.
Differs from prior one-shot talking face works, our method provides a new perspective to address this task and achieves vivid videos of arbitrary speakers.
The extensive quantitative and qualitative evaluations illustrate that our method is able to generate photo-realistic talking-face videos with accurate lip synchronization, natural lip shapes and rhythmic head motions from a reference image and a new audio clip.
Besides face photography, we can animate talking head videos for non-photorealistic paintings, demonstrating a promising generalization ability of our method. 

\bibliography{aaai22.bib}
\end{document}